\documentclass[nohyperref]{article}

\usepackage{microtype}
\usepackage{graphicx}
\usepackage{subfigure}
\usepackage{booktabs} 

\usepackage{hyperref}

\usepackage{makecell}

\usepackage[accepted]{icml2022}
\usepackage[symbol]{footmisc}

\usepackage{amsmath}
\usepackage{amssymb}
\usepackage{mathtools}
\usepackage{amsthm}
\usepackage{multirow}
\usepackage[capitalize,noabbrev]{cleveref}

\theoremstyle{plain}

\theoremstyle{definition}

\theoremstyle{remark}

\usepackage[textsize=tiny]{todonotes}

\icmltitlerunning{An Efficient Modern Baseline for FloodNet VQA}

\begin{document}

\twocolumn[
\icmltitle{An Efficient Modern Baseline for FloodNet VQA}



\icmlsetsymbol{equal}{*}
\begin{icmlauthorlist}
\icmlauthor{Aditya Kane}{equal,pict}
\icmlauthor{Sahil Khose}{equal,mit}
\end{icmlauthorlist}

\icmlaffiliation{pict}{Pune Institute of Computer Technology, Pune, India}
\icmlaffiliation{mit}{Manipal Institute of Technology, Manipal, India}

\icmlcorrespondingauthor{Aditya Kane}{adityakane1@gmail.com}
\icmlcorrespondingauthor{Sahil Khose}{sahilkhose18@gmail.com}

\icmlkeywords{Machine Learning, ICML}

\vskip 0.3in
]



\printAffiliationsAndNotice{\icmlEqualContribution} 

\begin{abstract}
Designing efficient and reliable VQA systems remains a challenging problem, more so in the case of disaster management and response systems. In this work, we revisit fundamental combination methods like concatenation, addition and element-wise multiplication with modern image and text feature abstraction models. We design a simple and efficient system which outperforms pre-existing methods on the FloodNet dataset and achieves state-of-the-art performance. This simplified system requires significantly less training and inference time than modern VQA architectures. We also study the performance of various backbones and report their consolidated results.\footnote[2]{Code is available at \href{https://bit.ly/floodnet_vqa_code}{https://bit.ly/floodnet\_vqa\_code}.}
\end{abstract}

\section{Introduction}
Floods remain one of the most common and large-scale impacting natural calamities. Millions of people are affected as a result of flooding each year. This can be averted by timely flood detection and urgent rescue operations. Firstly, quick data collection can be arranged by remote controlled drone systems which can collect high resolution images from flood affected regions. Following the data collection, a scene understanding and evaluation system is necessary to deploy task forces for quick evacuation. Visual Question Answering (VQA) system can prove to be of great use in such scenarios as it can provide answers to arbitrary questions encompassing many semantic understanding problems like object recognition, detection, attribute classification, counting, etc. Having reliable VQA systems for flood detection and understanding is vital as it will provide a rich variety of semantic information via the large number of scene understanding problems while also being the easily interpretable to the layman.

\citeauthor{vqaparikhetal} was the first seminal work to provide impactful results for the VQA task. Several works \cite{vlmo, ofa, florence, simvlm}  have further pushed the state-of-the-art in that direction. The FloodNet dataset addressed in this paper was introduced by \citeauthor{floodnet} The paper also provided several baselines for the 3 semantic understanding tasks -- classification, segmentation, and VQA. Even though a number of works address the problem of classification and segmentation \cite{semisupkhoseetal, chowdhury2021attention}, no work is available pertaining to VQA on the FloodNet dataset. Moreover, some baselines provided by \citeauthor{floodnet} are based on out-of-date models. This work has the following contributions:
\begin{enumerate}
    \item Provide an efficient method using modern image and text feature extraction architectures followed by basic combination methods achieving FloodNet VQA SOTA.
    \item Study the effects of various feature aggregation methods and compare them to well-known VQA models like ViLT \cite{vilt}.
\end{enumerate}

\begin{figure*}[!t]
\begin{center}
\includegraphics[width=\textwidth]{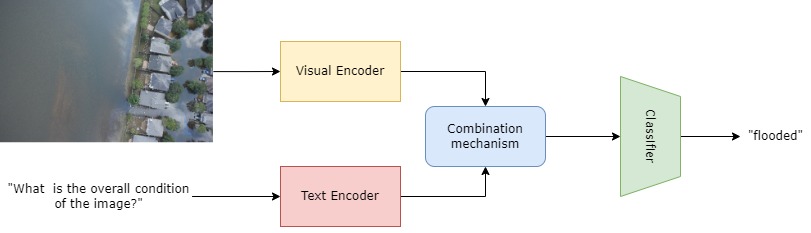}
\caption{Our simple VQA system}
\label{fig:floodnet_main}
\end{center}
\vskip -0.2in
\end{figure*}

\section{Dataset}
\begin{figure}[ht]
\begin{center}
\centerline{\includegraphics[width=\columnwidth]{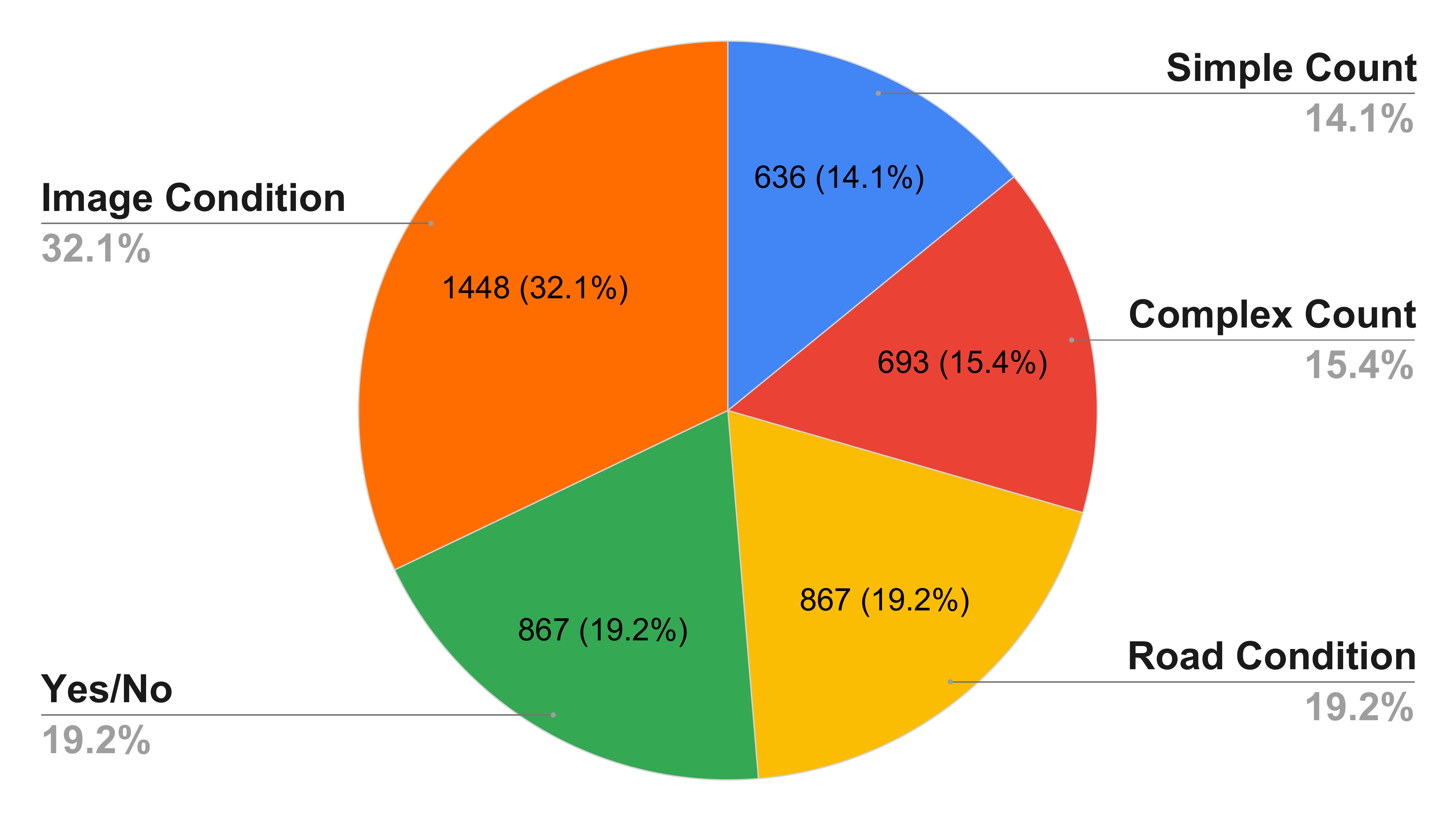}}
\caption{Distribution of types of questions}
\label{fig:dataset_split}
\end{center}
\vskip -0.2in
\end{figure}
To the best of our knowledge, FloodNet \cite{floodnet} is the first VQA work and dataset\footnote[3]{\href{https://bit.ly/floodnet_vqa_data}{https://bit.ly/floodnet\_vqa\_data}} focused on UAV imagery for any disaster damage assessment. We have access to the training labels hence we further split these into training and testing splits, since the ground truth labels of the original validation and testing splits are not available. We carefully bisect the available labelled data into training and testing data such that no images appear in both training and testing splits. Thus, our total usable dataset consists of 4511 question-answer pairs across 1448 images. Our training split consists of 3620 questions pertaining to 1158 images and our testing split has 891 questions pertaining to 290 images. Naturally, multiple question-answer pairs can be related to a single image. Each question is one of the following types: Simple Counting, Complex Counting, Condition of Road, Condition of Entire Image and Yes/No type of question. 
The dataset distribution is shown in Figure \ref{fig:dataset_split}.
These questions result in 56 probable labels, out of which the first five are represent non-numerical answers to the questions: ``flooded", ``non-flooded", ``flooded,non-flooded"  ``Yes" and ``No". The next fifty-one labels correspond to numerical answers, ranging from 0 to 50 (both inclusive).

\section{Proposed methods}

The methods which exhibit peak performance in FloodNet paper as well as the well-known ViLT \cite{vilt} models are compute expensive. They take long time to train and exhibit sub-optimal performance in some question types. We empirically show that simple methods like concatenation of features, element-wise multiplication or addition of visual and text features performs comparably or even outperforms the more compute intensive methods. It is noteworthy that we use features extracted from frozen pre-trained models, which are (1) easy to obtain and deploy in a real-world practical setting (2) quick to train and evaluate on new incoming data. Specifically, we use features extracted from RoBERTa-Large \cite{roberta} for extracting features from questions (text) and ResNet50 \cite{resnet} or ConvNeXt-Large \cite{convnext} for extracting features from images. Our general method is illustrated in Figure \ref{fig:floodnet_main}. We extract the image and text features beforehand from the aforementioned models and later use them to fine-tune rest of our model. This is computationally inexpensive as the large models are kept frozen and we only update a very small fraction of the total parameters. 

\subsection{Concatenation of features}

Our simplest approach consists of concatenating image and text features and passing them through three linear layers. Surprisingly, this approach has better results than MFB \cite{mfb} as reported in the FloodNet paper. 



\subsection{Element-wise addition or multiplication of features}

We convert the image and text features to a common dimension (here 512) using linear layers and take their element-wise summation or product. We pass this new feature vector through three linear layers to get our final prediction.

\section{Experiments and Results}
\begin{table*}[ht]
\scriptsize
  \centering
   
    \begin{tabular}{cccccccccc}
    \toprule
          & \multicolumn{1}{c}{} & \multicolumn{2}{c}{Time (mm:ss)} & \multicolumn{6}{c}{Accuracy (\%)} \\
    \midrule
          & \multicolumn{1}{c}{} &       &       &   \multicolumn{1}{c}{Overall}    & \multicolumn{2}{c}{Counting Problem} &        \multicolumn{3}{c}{Condition Recognition} \\
    \midrule
    
    
    \multicolumn{1}{c}{Group} & Method & \multicolumn{1}{c}{Training} & \multicolumn{1}{c}{Inference} & \multicolumn{1}{c}{} & \multicolumn{1}{c}{Simple Count} & \multicolumn{1}{c}{Complex Count} & \multicolumn{1}{c}{Yes/No} & \multicolumn{1}{c}{Image Condition} & \multicolumn{1}{c}{Road Condition} \\
    \midrule
    \multicolumn{1}{c}{\multirow{4}[8]{*}{Floodnet baselines}} & base-cat & \multicolumn{1}{c}{-} & \multicolumn{1}{c}{-} & 42    & 4     & 3     & 17    & 86    & 90 \\
\cmidrule{2-10}          & base-mul & \multicolumn{1}{c}{-} & \multicolumn{1}{c}{-} & 68    & 25    & 21    & 84    & 96    & 97 \\
\cmidrule{2-10}          & SAN   & \multicolumn{1}{c}{-} & \multicolumn{1}{c}{-} & 63    & 26    & 24    & 54    & 94    & 97 \\
\cmidrule{2-10}          & MFB   & \multicolumn{1}{c}{-} & \multicolumn{1}{c}{-} & 73    & 29    & 26    & 99    & 97    & \textbf{99} \\
    \midrule
    \multicolumn{1}{c}{\multirow{11}[22]{*}{Our contributions}} & ViLT@1 & 24:00 & 6:00 & 73.73 & 22.41 & 20.61 & 86.44 & 96.55 & 96.61 \\
\cmidrule{2-10}          & ViLT@2 & 48:00 & 6:00 & 76.43 & 24.14 & 20.61 & 98.87 & 96.55 & 96.61 \\
\cmidrule{2-10}          & ViLT@3 & 72:00 & 6:00 & 76.43 & 25.86 & 20.61 & \textbf{99.43} & 95.51 & 96.61 \\
\cmidrule{2-10}          & ViLT@4 & 96:00 & 6:00 & 76.09 & 26.72 & 19.08 & \textbf{99.43} & 94.82 & 96.61 \\
\cmidrule{2-10}          & ViLT@5 & 120:00 & 6:00 & 74.97 & 26.72 & 19.08 & \textbf{99.43} & 91.38 & 96.61 \\
\cmidrule{2-10}          & R50-cat & 15:24 & 3:32  & 77.41 & 34.48 & 33.46 & 82.76 & 98.7  & \textbf{97.42} \\
\cmidrule{2-10}          & R50-mul & 17:24 & 3:32  & 80.36 & 35.34 & 32.06 & 98.87 & \textbf{98.62} & 97.18 \\
\cmidrule{2-10}          & R50-add & 17:24 & 3:32  & 77.55 & 36.21 & 32.82 & 83.62 & \textbf{98.62} & 97.15 \\
\cmidrule{2-10}          & CNX-cat & 7:16  & 1:24  & 74.41 & \textbf{40.51} & 35.11 & 63.84 & 98.28 & 97.18 \\
\cmidrule{2-10}          & \textbf{CNX-mul} & 9:16  & 1:24  & \textbf{81.26} & 37.07 & \textbf{37.4}  & 98.31 & \textbf{98.62} & 97.18 \\
\cmidrule{2-10}          & CNX-add & 9:16  & 1:24  & 72.05 & 37.93 & 34.35 & 54.8  & 98.28 & 96.61 \\
    \bottomrule

    \end{tabular}%
  \caption{Our results on FloodNet dataset}
  \label{tab:results}%

\end{table*}%

We performed extensive experiments using various combination methods and two vision encoder models: ResNet50 and ConvNeXt. We used RoBERTa-Large for encoding question text. We froze the model layers and used the pre-trained encoders for image and text features. This resulted in drastic reduction in training and evaluation time compared to VQA models like ViLT. Our training setup was on the Google Colab platform. For our model training and inference, we had access to NVIDIA Tesla T4 GPU with 15GB VRAM attached to a 2-core Xeon CPU with 12GB RAM. We use CrossEntropy loss for our training. The simpler combination methods used Adam with a LR of $3e-4$ and a batch size of 128 for 100 epochs. For ViLT, we used AdamW with a LR of $5e-5$ and batch size of $1$ for 5 epochs.

Our complete results are available in Table \ref{tab:results}. We use the shorthand ``(model)-(method)" for the sake of brevity, where ``model" can be one of ``base" (baseline from the FloodNet paper), ``R50" (ResNet50) or ``CNX" (ConvNeXt) and ``method" can be one of ``cat" (concatenation), ``add" (addition), ``mul" (element wise multiplication). ``ViLT@k" denotes the ViLT model trained for ``k" epochs. We present a few observations from the results below:

\begin{enumerate}

    \item \textbf{Modern text encoders aid VQA}: Modern transformer-based architectures like RoBERTa outperform LSTMs on all of the current language benchmarks. This dominant performance indicates that they have a richer semantic understanding of language. As a result of these semantically better representations, our VQA performance seems to increase dramatically than their LSTM-based counterparts for similar methods.
    
    \item \textbf{Modern image encoders aid VQA}: Both ResNet50 and ConvNeXt backbone methods outperform previous methods. We observe major improvements in overall accuracy, primarily stemming from the boost in Counting Problem performance compared to the FloodNet and ViLT results. ConvNeXt results achieve the best overall metrics predominantly due to the best Counting Problem performance on the table, even after taking a minor hit to ``Yes/No" accuracy compared to its ResNet50 counterpart. Counting Problem is one of the most challenging tasks as it requires a deeper scene understanding coupled with determining the answer between 51 labels out of the 56 total for this VQA task. 
    
    \item \textbf{Feature multiplication is the best combination method}:
    Feature multiplication using both ResNet50 and ConvNeXt outperforms other combination methods (concatenate, add) that we explore. We speculate that this  leads to direct interaction of image and text features resulting in the gradients of the two modalities being tightly coupled with each other leading to a better common semantic space essential for VQA. 
    
    \item \textbf{ViLT is good, but not the best}: 
    ViLT is a transformer-based architecture that processes images in a patch-wise manner. Thus, it is computationally expensive and requires a longer training time to converge. Although with just one epoch of training ViLT is able to outperform previous FloodNet benchmarks, it performs poorly compared to our simpler combination methods. It achieves peak metrics at epoch two but our best setting achieves almost 5\% better performance with five times less training time. Due to computation constraints, we report results on five epochs of ViLT.

\end{enumerate}

\section{Conclusion and Future Work}
In this work, we explore the VQA task on the FloodNet dataset intending to facilitate quick training and better performance for rapid flood rescue response. We revisit simpler combination mechanisms for VQA with modern architectures for vision and language encoders. Our best models achieve SOTA performance on the FloodNet dataset for the VQA task with significantly less time for training and inference than the modern VQA architecture (ViLT). We foresee multiple opportunities for future research. A deployable system can be strived for by compressing the models using various distillation and pruning methods. Such a disaster system would also benefit from a continual learning setup that reflects a more realistic scenario for our use case.

\bibliography{paper}
\bibliographystyle{icml2022}



\end{document}